\documentclass[sigconf, nonacm]{acmart}
\usepackage{algorithm}
\usepackage{algpseudocode}

\usepackage{multirow}
\usepackage{diagbox}
\usepackage{graphicx}
\usepackage{subcaption}  % For creating subfigures
\usepackage{threeparttable}
\usepackage{booktabs} % 用于绘制更美观的横线
\usepackage{enumitem}
\newcommand\tool{\textsc{SAP-DIFF}}

\usepackage{ulem}

\begin{document}

\title{\tool: Semantic Adversarial Patch Generation for Black-Box Face Recognition Models via Diffusion Models}
\author{Anonymous authors}

\author{Mingsi Wang}
\email{wangmingsi@iie.ac.cn}
\affiliation{%
  \institution{Institute of Information Engineering, Chinese Academy of Sciences}
  \institution{School of Cyber Security, University of Chinese Academy of Sciences}
  \city{Beijing}
  \country{China}
}

\author{Shuaiyin Yao}
\email{2574623613ysy@gmail.com}
\affiliation{%
  \institution{Institute of Information Engineering, Chinese Academy of Sciences}
  \institution{School of Cyber Security, University of Chinese Academy of Sciences}
  \city{Beijing}
  \country{China}
}

\author{Chang Yue}
\email{yuechang@iie.ac.cn}
\affiliation{%
  \institution{Institute of Information Engineering, Chinese Academy of Sciences}
  \institution{School of Cyber Security, University of Chinese Academy of Sciences}
  \city{Beijing}
  \country{China}
}

\author{Lijie Zhang}
\email{zhanglijie@iie.ac.cn}
\affiliation{%
  \institution{Institute of Information Engineering, Chinese Academy of Sciences}
  \institution{School of Cyber Security, University of Chinese Academy of Sciences}
  \city{Beijing}
  \country{China}
}

\author{Guozhu Meng}
\authornote{Corresponding author}
\email{mengguozhu@iie.ac.cn}
\authornotemark[1]
\affiliation{%
  \institution{Institute of Information Engineering, Chinese Academy of Sciences}
  \institution{School of Cyber Security, University of Chinese Academy of Sciences}
  \city{Beijing}
  \country{China}
}

\begin{abstract}
  % Face recognition models have been widely applied in finance, military, public security, and everyday life.
  Given the need to evaluate the robustness of face recognition (FR) models, many efforts have focused on adversarial patch attacks that mislead FR models by introducing localized perturbations.
  % Impersonation attacks represent a significant threat resulting from adversarial perturbations, where attackers can disguise themselves as legitimate users to perform unauthorized actions, potentially resulting in severe consequences such as data breaches, system damage, and resource misuse. 
  Impersonation attacks are a significant threat because adversarial perturbations allow attackers to disguise themselves as legitimate users. This can lead to severe consequences, including data breaches, system damage, and misuse of resources.
  However, research on such attacks in FR remains limited.
  Existing adversarial patch generation methods exhibit limited efficacy in impersonation attacks due to (1) the need for high attacker capabilities, (2) low attack success rates, and (3) excessive query requirements.
  To address these challenges, we propose a novel method \tool~ that leverages diffusion models to generate adversarial patches via semantic perturbations in the latent space rather than direct pixel manipulation.
  We introduce an attention disruption mechanism to generate features unrelated to the original face, facilitating the creation of adversarial samples and a directional loss function to guide perturbations toward the target identity’s feature space, thereby enhancing attack effectiveness and efficiency.
  Extensive experiments on popular FR models and datasets demonstrate that our method outperforms state-of-the-art approaches, achieving an average attack success rate improvement of 45.66\% (all exceeding 40\%), and a reduction in the number of queries by about 40\% compared to the SOTA approach.
\end{abstract}






\maketitle

\section{Introduction}
\label{sec:introduction}
Face Recognition (FR) models have significantly advanced in accuracy and scalability, largely due to breakthroughs in deep learning~\cite{lecun2015deep, DeepFace, schroff2015facenet}. Consequently, these models have become fundamental in diverse applications, including device unlocking~\cite{patel2016secure}, e-banking~\cite{wang2021exploring}, and military operations~\cite{goel2019development}.
FR models typically employ highly trained models that analyze facial features extracted from face images, allowing for effective identification and authentication of individuals. However, security concerns have escalated with the rapid growth of FR models. 
Recent research~\cite{szegedy2013intriguing, dong2019efficient, zhong2020towards,wang2024survey,brown2017adversarial}  has demonstrated that FR models are vulnerable to adversarial attacks, where maliciously designed samples are used to bypass the security of FR models, leading to misidentification of individuals. 

Global perturbation is a type of adversarial attack that modifies the pixel values across the entire face image. The magnitude of the perturbation is often constrained by the Lp-norm to ensure imperceptibility to human observers. However, such global perturbations are generally confined to the digital domain and are not practical~\cite{wang2024survey,brown2017adversarial}. To address this limitation, Brown \textit{et al.}~\cite{brown2017adversarial} introduced the concept of adversarial patches, which place a patch in a specific region of the face.
Several approaches have been developed to realize adversarial patches. 
For instance, some methods utilize adversarial wearable accessories into such as hats with adversarial stickers~\cite{komkov2021advhat}, glasses embed perturbations within custom-designed eyeglass frames~\cite{sharif2019general, sharif2016accessorize} or even adversarial makeup applied to facial regions~\cite{an2023imu, yin2021adv}. Others employ light projection onto a person's face~\cite{wang2024invisible, yamada2012use, shen2019vla} to introduce dynamic adversarial patterns.

Adversarial patch attack can result in either a dodging attack (the FR system fails to identify the attacker's identity correctly) or an impersonation attack (the FR system incorrectly identifies the attacker as another legitimate user)~\cite{wang2024survey}.
Among these, impersonation attacks pose a more severe threat, as they allow attackers to forge a legitimate identity and execute unauthorized actions. Unlike dodging attacks, which primarily affect authentication failures, impersonation attacks directly undermine identity verification mechanisms, potentially facilitating fraudulent transactions, unauthorized access, and privacy violations. Given these high-stakes risks, research on adversarial impersonation attacks is crucial for improving the robustness and security of face recognition systems~\cite{wang2022survey, wei2022visually, guesmi2023physical, wang2023adversarial}.

However, previous impersonation attack methods often face several limitations: (1) High attacker capability requirements. Many existing approaches rely on white-box models, assuming access to model parameters, intermediate feature representations, and gradients~\cite{zheng2023robust, gong2023stealthy, zolfi2022adversarial}, which places excessively high demands on the attacker's capabilities, significantly limiting the practical applicability; (2) Limited attack success rate. Previous methods often fail to exploit the semantic information inherent in face images, limiting their effectiveness to impersonation attacks; and (3) Excessive query overhead. Some methods require an extensive number of queries to the target model to iteratively optimize the adversarial patches~\cite{wei2024physical, fang2024state, kong2022digital}. This not only increases computational costs but also raises concerns about practical deployment.

To address the above issues, we propose a novel query-based black-box impersonation attack leveraging diffusion models. The diffusion model has been proven to be highly effective in capturing and extracting semantic features from images~\cite{ho2020denoising, song2021denoising}, enabling more precise control over the patch's semantics, thereby achieving better attack performance and generation efficiency. Specifically, we initialize the adversarial patch with DDIM Inversion~\cite{song2021denoising} as a latent representation. Then, during each query round, we optimize this latent representation in the diffusion model’s denoising process.

In each epoch of optimization, we perform the following steps:
First, we disrupt the attention mechanism of the diffusion model, preventing it from generating features related to the benign face image. This ensures that the generated patch is effectively separated from the original face's features, which is necessary to achieve a successful impersonation attack by avoiding unwanted blending with the original face's characteristics.
Second, we directionally guide the patch's features toward the target identity’s feature space, aligning them with the desired target face. These two operations are performed on the latent obtained after one step of denoising.
Finally, we decode the optimized latent embedding to pixel space and use the UV location map~\cite{feng2018joint} to realistically place the generated adversarial patches on the face. We then adjust the patches based on the target model's output to ensure the adversarial samples successfully attack the target model. These three steps together complete a full epoch of optimization.

\tool~ allows us to bypass the need for explicit access to the target model’s parameters, instead relying on queries to assess the model’s responses and iteratively refine our adversarial patches. In addition, by using an additional diffusion model for adversarial patch semantic feature extraction and optimization direction guidance, we reduce the dependency on queries to the target model.

Extensive experiments on benchmark datasets (LFW, CelebA-HQ) across popular FR models (ArcFace, CosFace, FaceNet) demonstrate that \tool~ outperforms state-of-the-art approaches. In all tasks, the attack success rate is improved by an average of 45.66\% (all improved above 40\%), and a reduction in the number of queries by about 40\% compared to the SOTA approach. 

The main contributions of our paper are summarized as follows:
\begin{itemize} 
    \item We propose a novel adversarial patch generation framework \tool~ that leverages latent space manipulation through a diffusion model, enabling the creation of semantically effective adversarial patches for query-based black-box impersonation attacks. 
    \item We design an optimization strategy that integrates attention disruption, directional loss, and UV location map, ensuring the generation of adversarial patches that effectively deceive FR models while maintaining semantic integrity. 
    \item We conduct extensive evaluations on popular FR models and datasets, demonstrating that \tool~ achieves superior attack success rate and query efficiency compared to existing state-of-the-art adversarial patch attacks.
\end{itemize}

\section{Related Work and Background}
\label{sec:related work} 
\subsection{Adversarial Attacks for Face Recognition}
Face recognition models have been demonstrated to be vulnerable to adversarial attacks~\cite{szegedy2013intriguing, dong2019efficient, zhong2020towards, wang2024survey}. These attacks can be broadly classified into two categories: (a) global perturbations, which involve directly modifying pixel values in the entire face image, resulting in significant changes to the overall appearance and causing the model to misidentify the input face image; and (b) localized perturbations, which generate adversarial patches that are applied to specific regions of the face image, strategically inducing misidentification without altering the entire image.

\noindent\textbf{Global Attacks.} Numerous studies \cite{dong2019efficient, zhong2020towards, yang2021attacks, deb2020advfaces, jia2022adv, goswami2018unravelling, zhong2019adversarial, chatzikyriakidis2019adversarial, dabouei2019fast} have proposed methods for generating subtle yet potent global perturbations that modify the entire face image. These perturbations target every pixel in the image, making it difficult for the human eye to detect significant changes while significantly reducing the performance of face recognition systems. 
Recently, diffusion models have emerged as an alternative approach for generating global adversarial attacks~\cite{couairon2023diffedit, chen2023advdiffuser, sun2024diffam, chen2024diffusion}. Unlike traditional gradient-based methods, diffusion models can introduce semantic, yet subtle perturbations that are difficult to detect by both human observers and recognition systems. This makes them an appealing option for generating global attacks. Despite the subtlety and imperceptibility of these perturbations, they exploit vulnerabilities within face recognition models by altering the pixel values throughout the image. 
However, global perturbations are often vulnerable to countermeasures such as adversarial training and purification techniques, which can significantly mitigate their effectiveness. Additionally, their practical applicability is limited, as they require full image-level modifications that are not easily transferable to practical scenarios.

\noindent\textbf{Patch Attacks.} Compared to pixel-wise imperceptible global perturbations, adversarial patches do not restrict the magnitude of perturbations. 
Attackers have proposed various techniques to introduce localized patches to face recognition models. For example, adversarial hat~\cite{komkov2021advhat}, adversarial mask~\cite{xiao2021improving, yang2022controllable, yang2023towards}, adversarial sticker~\cite{wei2022adversarial, wei2022simultaneously, pautov2019adversarial, xiao2021improving} and adversarial glasses~\cite{sharif2019general, sharif2016accessorize} are classical methods against face recognition models which are realized by placing perturbation patches on the forehead or nose or putting the perturbation eyeglasses on the eyes. GenAP~\cite{xiao2021improving} optimizes the adversarial patch on a low dimensional manifold and pastes them on the area of eyes and eyebrows. Face3DAdv~\cite{yang2022controllable} leverages a 3D generator to synthesize face information and introduces a texture-based adversarial attack to render the patch into 2D faces. AT3D~\cite{yang2023towards} introduces adversarial textured 3D meshes with elaborate topology for adversarial patch attacks, utilizing low-dimensional coefficient perturbations based on the 3D Morphable Model to enhance black-box transferability. While these transfer-based methods improve the adaptability of adversarial patches to different recognition models, their effectiveness is still limited, with lower success rates in impersonate attacks due to insufficient transferability.
RHDE~\cite{wei2022adversarial} utilizes a pattern-fixed sticker existing in real life to attack black-box FR systems by querying patch locations through the differential evolution algorithm.
Wei \textit{et al.}~\cite{wei2022simultaneously} utilize reinforcement learning to simultaneously solve the optimal solution for the patch location and perturbation through queries based on the rewards obtained from the target model. However, these methods suffer from low query efficiency, limiting their practical effectiveness.
Our work shows how to adequately use diffusion models to improve the effectiveness of adversarial patches and query efficiency.

% \vspace{-5pt}

\subsection{Diffusion Models}
Diffusion models have recently garnered considerable attention in the machine learning community primarily due to their impressive capability to generate high-quality samples by effectively modeling data distributions through iterative denoising processes. As mentioned in~\cite{chen2024diffusion}, two fundamental approaches within this family are the Denoising Diffusion Probabilistic Model (DDPM)~\cite{ho2020denoising} and the Denoising Diffusion Implicit Model (DDIM)~\cite{song2021denoising}. DDIM is the foundation of the method employed in our \tool.

DDPMs are generative models that formulate the data generation process as a Markovian chain of Gaussian transitions.
The model consists of a forward process, which progressively corrupts data into pure noise, and a reverse process, which learns to recover the original data from noise. 

In the forward process, data ${x}_0$ is progressively transformed into latent variables ${x}_1, {x}_2, \ldots, {x}_T$ over $T$ timesteps through a sequence of Gaussian noise injections:
\begin{equation}
q({x}_t | {x}_{t-1}) = \mathcal{N}({x}_t; \sqrt{1 - \beta_t} {x}_{t-1}, \beta_t {I}),
\end{equation}

where $\beta_t$ is the noise schedule. This allows direct sampling of ${x}_t$ from ${x}_0$ using:
\begin{equation}
q({x}_t | {x}_0) = \mathcal{N}({x}_t; \sqrt{\bar{\alpha}_t} {x}_0, (1 - \bar{\alpha}_t) {I}),
\end{equation}
\begin{equation}
x_t=\sqrt{\bar{\alpha}_t}x_0+\sqrt{1-\bar{\alpha}_t}\epsilon,\quad\epsilon\sim\mathcal{N}(0,{I}),
\end{equation}
where $\alpha_t=1-\beta_t$. $\bar{\alpha}_t=\prod_{s=0}^t\alpha_s$ is the cumulative product of the noise schedule. This allows efficient sampling of noisy data without iteratively applying all intermediate steps.

The reverse process aims to denoise ${x}_t$ step by step, reconstructing ${x}_0$. Since the true posterior $q({x}_{t-1} | {x}_t)$ is intractable, a neural network $p_\theta$ is used to approximate it as~\cite{sohl2015deep}:
\begin{equation}
p_\theta({x}_{t-1} | {x}_t) = \mathcal{N}({x}_{t-1}; \mu_\theta({x}_t, t). \Sigma_\theta({x}_t, t)),
\end{equation}
where \( \mu_\theta(x_t, t) \) is the predicted mean. The objective is to minimize the error between predicted and true noise~\cite{ho2020denoising}:
\begin{equation}
\label{eq: noise}
\mathcal{L}(\theta) = \mathbb{E}_{x_0, \epsilon, t} [ \| \epsilon - \epsilon_\theta(x_t, t) \|^2 ].
\end{equation}
Sampling is then done with the trained $\theta({x}_t, t)$ :
\begin{equation}
x_{t-1}=\mu_\theta\left(x_t, t\right)+\sigma_t z, \quad z \sim N(0, I)
\end{equation}

DDPM employs a stochastic reverse process, whereas DDIM introduces a deterministic alternative to enhance sampling efficiency while still maintaining sample quality. 
Instead of a Markovian process, DDIMs use a fixed transformation to map noise to data~\cite{song2021denoising}:
\begin{equation}
{x}_{t-1} = \sqrt{\alpha_{t-1}} \left( \frac{{x}_t - \sqrt{1 - \alpha_t} \epsilon_\theta({x}_t, t)}{\sqrt{\alpha_t}} \right) + \sqrt{1 - \alpha_{t-1}} \epsilon_\theta({x}_t, t),
\end{equation}
where $\alpha_t$ is a cumulative product of $(1 - \beta_t)$ over timesteps, and $\epsilon_\theta$ represents the predicted noise.

The deterministic nature of DDIM can be interpreted as Euler integration for solving ordinary differential equations (ODEs)~\cite{song2021denoising}. This perspective enables the reverse process to effectively map a real image to its corresponding latent representation, facilitating a more stable transformation. This operation referred to as DDIM Inversion, facilitates subsequent manipulations of real images~\cite{song2021denoising}. Conversely, the reverse process in DDIM can be expressed as:
\begin{equation}
    \begin{aligned}x_{t+1}-x_{t}=&\sqrt{\bar{\alpha}_{t+1}}\left[\left(\sqrt{1/\bar{\alpha}_{t}}-\sqrt{1/\bar{\alpha}_{t+1}}\right)x_{t}\right.\\&+\left(\sqrt{1/\bar{\alpha}_{t+1}-1}-\sqrt{1/\bar{\alpha}_{t}-1}\right)\epsilon_{\theta}(x_{t},t)\Big]\end{aligned}
\end{equation}

$x_t$ is the output of $x$ after $t$ timesteps of DDIM Inversion, denoted as $DDIM_{inverse}(x, t)$.

The denoising process of DDIM (denoted as ${DDIM_{denoise}}$) operates as follows:
\begin{equation}
    \begin{aligned}x_{t-1}-x_{t}=&\sqrt{\bar{\alpha}_{t-1}}\left[\left(\sqrt{1/\bar{\alpha}_{t}}-\sqrt{1/\bar{\alpha}_{t-1}}\right)x_{t}\right.\\&\left.+\left(\sqrt{1/\bar{\alpha}_{t-1}-1}-\sqrt{1/\bar{\alpha}_{t}-1}\right)\epsilon_{\theta}(x_{t},t)\right]\end{aligned}
\end{equation}

$x_t$ is the output of the DDIM denoising process  after $t$ timesteps applied to $x$, denoted as $DDIM_{denoise}(x, t)$.

This framework provides a principled approach for both denoising and latent encoding, paving the way for diverse applications in real-image manipulation and generation~\cite{mokady2023null}~\cite{couairon2023diffedit}.

\section{Methodology}
\label{sec:method}
This section introduces our method \tool, which generates adversarial patches on face recognition models with diffusion models. First, we formulate our problem: generating adversarial patches to perform impersonation attacks against face recognition models (Sec.~\ref{subsec: Problem Formulation}), then present our solution (Sec.~\ref{subsec: Basic Framework}), and finally give the details of our approach (Sec.~\ref{subsec: Patch Initialization} and Sec.~\ref{subsec: Patch Optimization}).

\subsection{Problem Formulation}
\label{subsec: Problem Formulation}
We aim to implement impersonation attacks against face recognition models. Given a clean face image $x$ and a target face image $x^{tar}$, our goal is to generate an adversarial patch that makes the face recognition model $f_\theta$ ($\theta$ denotes the model's parameters) predict the target identity for the perturbed face image $x^{adv}$. Formally, the goal is to generate the patch $p$ such that:
\begin{equation}
    f_{\theta}(x\odot p)= f_{\theta}(x^{adv}) \approx f_{\theta}(x^{tar}),
\end{equation}
where $p$ is the adversarial patch (adversarial mask in our method) and $\odot$ denotes the operation of applying the patch to the image $x$. 

We operate under the assumption that no prior knowledge of the target model's specific parameters, architecture, or training data is available. Instead, we conduct query-based black-box attacks, relying solely on the ability to query the model and retrieve its output embeddings. To incorporate the semantic features of the target face into the adversarial patch, we optimize the patch $p$ within the latent space of the diffusion model, leveraging its properties to generate semantic and effective attacks. In the following sections, we provide a detailed explanation of our design.

% \subsection{Basic Framework our solution}
\subsection{Our Solution}
\label{subsec: Basic Framework}
\begin{figure*}[!t]
\centering
\includegraphics[width=0.90\textwidth]{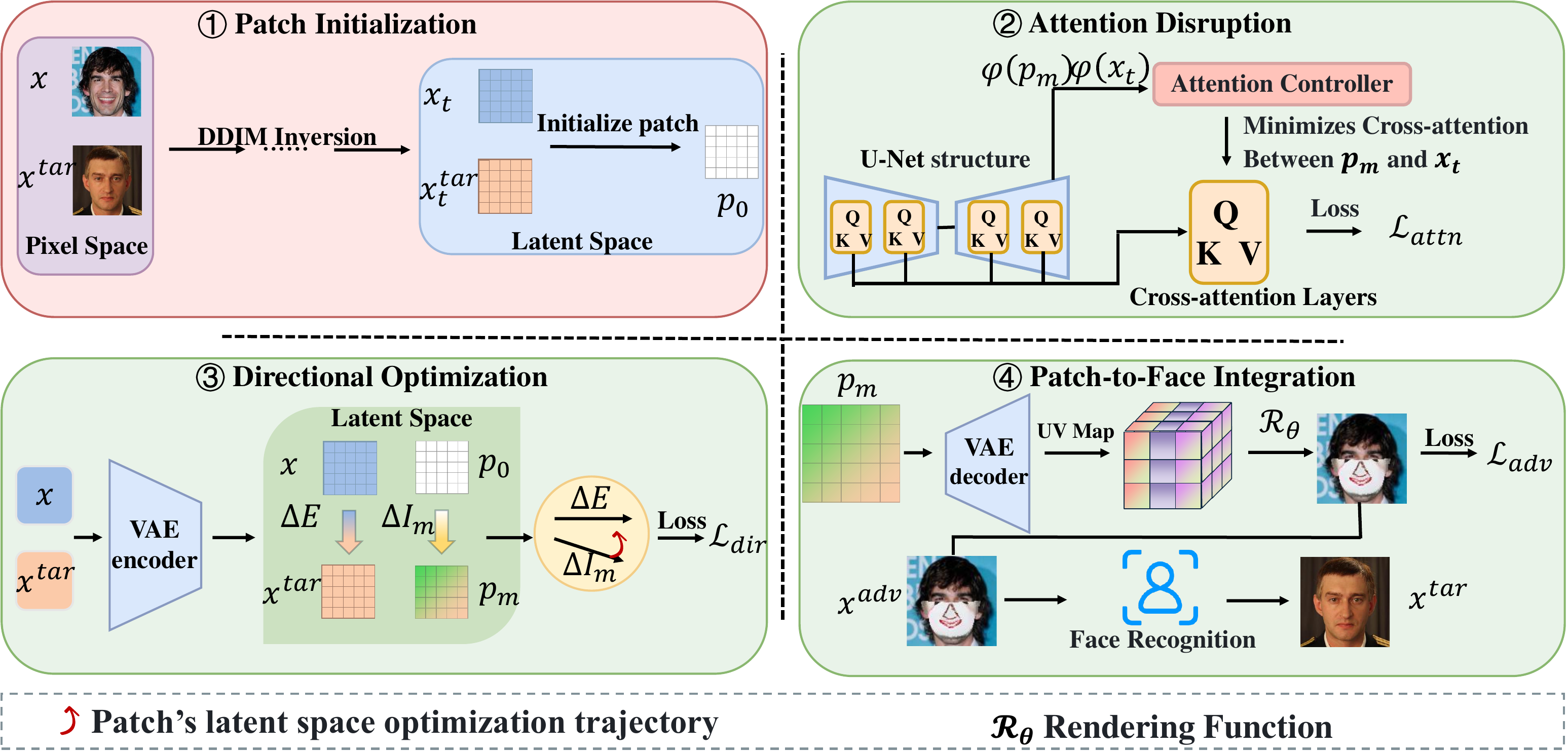}
\caption{Overview of \tool.  Here, we present the four components of our method, where both $\mathcal{L}{attn}$ and $\mathcal{L}{dir}$ are computed in the latent space and then transferred to the image domain for patch placement, contributing to the calculation of $\mathcal{L}_{adv}$.}
\label{fig: overview}
\end{figure*}

The framework of \tool~is shown in Fig.~\ref{fig: overview}, comprises two key stages: patch initialization and patch optimization.

\noindent\textbf{Patch Initialization.} We begin in the latent space by combining the original and target faces' latent embeddings to form the adversarial patch's initial embedding. This weighted combination incorporates the target features while maintaining the imperceptibility of the patch. 
Unlike pixel-level perturbations in the image domain, which are prone to noise and lack semantic alignment, the latent space leverages the rich semantic representations of the diffusion model, enabling structured and interpretable modifications that ensure meaningful and effective adversarial edits~\cite{chen2023advdiffuser}.
Specifically, we leverage DDIM Inversion to map the image into the latent space, enabling precise fusion of clean and target latent representations. This process guarantees that the adversarial patch remains realistic and maintains high-quality reconstructions when mapped back to the image domain. Additionally, latent space manipulation enhances the universality of adversarial perturbations across images by aligning with the underlying semantic structure rather than superficial pixel-based differences~\cite{sun2024diffam, couairon2023diffedit}. 

\noindent\textbf{Patch Optimization.} To optimize the patch for the impersonation attack, which is to make the face recognition model classify the adversarial sample as the target identity, we design three loss functions to address different optimization objectives.

\textit{Attention Disruption Loss.} In the latent space, we employ an attention disruption mechanism to interfere with the diffusion model's attention, preventing the generation of features related to the benign face image during the denoising optimization process. This mechanism aims to push the adversarial patch's features away from the original image's characteristics, which is essential for avoiding unwanted blending and ensuring a successful impersonation attack.

\textit{Directional Loss.} This loss aligns the adversarial patch’s latent-space perturbations with the semantic direction from the source to target face features, ensuring the patch actively injects the target’s attributes rather than merely deviating from the source. Enforcing this directional consistency improves the success rate of impersonation through targeted feature-space manipulation.

\textit{Attack Loss.} After decoding the patch’s latent embedding into a patch image, we use a UV location map~\cite{feng2018joint} to digitally place the patch onto the face, creating the adversarial sample. To ensure the adversarial patch achieves the impersonation goal, we compute a cosine similarity loss between the adversarial sample and the target face image. This ensures the adversarial patch's effectiveness in fooling the face recognition model.

\subsection{Patch Initialization}
\label{subsec: Patch Initialization}
We adopt open-source stable diffusion~\cite{rombach2022high} that is pre-trained on extremely massive text-image pairs. Since adversarial patch attacks aiming to fool the target model can be approximated as a special kind of real image editing, \tool~utilize the DDIM Inversion technology~\cite{song2021denoising}. The image is mapped back into the diffusion latent space by reversing the deterministic sampling process, thereby enabling precise control over the image features during reconstruction. The patch is initialized as follows:
% \begin{equation}p_t={DDIM_{inverse}}((1- \alpha) x_{t-1}
% +\alpha x^{tar}_{t-1})\end{equation}

\begin{equation}p_0= (1- \alpha) x_{t} + \alpha  x^{tar}_{t},
\end{equation}
where $p_0$ is the initialized patch latent embedding, $t$ is set as 5~\cite{chen2024diffusion}.

Specifically, the latent representations \( x_{t} \) and \( x^{tar}_{t} \) are generated by applying the DDIM Inversion to the initial representations \( x \) and \( x^{tar} \) for $t$ timestep, $x_{t} =DDIM_{inverse}(x, t)$, $x^{tar}_{t} =DDIM_{inverse}(x^{tar}, t)$. The latent representations of the clean image and the target image are combined by applying a weight $\alpha$, which controls the relative contributions of the original and target features. 
By progressively performing this inversion, we generate latent representations $x_{t}$ and $x^{tar}_{t}$ that incorporate the benign and target features. Through this process, the adversarial patch can be reconstructed with high quality, ensuring that the perturbations are both realistic and effective at deceiving the model.

\subsection{Patch Optimization}
\label{subsec: Patch Optimization}
\noindent\textbf{Attention Disruption.} 
The attention mechanism plays a crucial role in controlling which parts of the input features are emphasized during the image generation process. By intervening in this mechanism, we can guide the diffusion model to generate adversarial patches that are distant from the original face features in the latent space, ensuring the patches' effectiveness while minimizing their similarity to benign features.

To achieve this, \tool~leverage the attention mechanism within diffusion models to control the generation of adversarial patches for FR models effectively. 
Specifically, we incorporate the original latent $x_t$ to represent the source face feature. During the diffusion model's denoising process, we manipulate the cross-attention layers by using an attention controller. This controller adjusts the attention values dynamically to control the interaction between the adversarial patch and the original face features.

We first obtain the deep features, denoted as \(\varphi({p}_m)\), from the latent embedding of the patch, and the deep features of the original face, \(\varphi(x_t)\), within the U-Net structure. Here, $\varphi$ represents the operation within the U-Net architecture that extracts hierarchical and multi-scale features, and $m$ denotes the optimization epoch. These features are then projected into $Q$ and $K$ matrices using linear projections with matrices $W_Q $ and $W_K$. The fusion of the patch information proceeds as follows:

\begin{equation}
\text{softmax}\left(\frac{QK^T}{\sqrt{d}}\right) = \text{softmax}\left(\frac{\left(\varphi({p}_m) W_Q\right)\left(\varphi(x_t) W_K\right)^T}{\sqrt{d}}\right),
\end{equation}
where d denotes the dimension of $W_Q $ and $W_K$.

This manipulation ensures that we can influence the attention mechanism to emphasize the original facial features less. To quantify this effect, we use the following loss function:
\begin{equation}
\mathcal{L}_{attn} = \sigma^2 \left( \mathbb{E} \left( \mathcal{C}(p_m, x_t; \text{SD}) \right) \right),
\end{equation}
where $\sigma^2$ denotes the input’s variance, $\mathcal{C}$ denotes the cross-attention values in the denoising process, and SD is the Stable Diffusion. This loss function minimizes the cross-attention between the adversarial patch and the original face, effectively breaking the semantic connections between the two in the generated image. By dynamically adjusting the attention values throughout the patch generation process, we ensure that the diffusion model generates adversarial patches that are spatially and semantically distinct from the original face, thus enhancing the success of the impersonation attack.

\noindent\textbf{Directional Optimization.} In the context of an impersonation attack, merely disrupting the attention mechanisms is not enough to guide the adversarial patch toward the desired target. To effectively steer the perturbations toward the specific identity of the target, we extend the attention manipulation approach by introducing a direction loss. This direction loss aims to align the adversarial patch's optimization process with the target face’s feature space, ensuring that the generated patch not only deviates from the source image but also resembles the target's facial attributes.

To achieve this, we leverage the Variational Autoencoder (VAE), which is an integral part of the Stable Diffusion model. The VAE is responsible for encoding both the source and target images into a shared latent space. This latent representation captures the essential, semantically meaningful features of the face images, enabling the adversarial patch to interact with the diffusion model in a controlled manner~\cite{esser2021taming}. While the VAE is a standard component of Stable Diffusion, we utilize it to guide the generation process within the latent space, where the diffusion model iteratively refines the patch generation based on the encoded information.

To guide the optimization process toward the desired features, we design the direction loss to align the adversarial patch’s latent space optimization trajectory with the feature direction defined by the difference between the target and source face encodings:
\begin{equation}
    \mathcal{L}_{dir}=1-\frac{\Delta I_m\cdot\Delta E}{\|\Delta I_m\|\|\Delta E\|},
\end{equation}

where $\Delta I_{m}=p_{m} - p_{0}$, $\Delta E =\mathcal{E}(x^{tar})-\mathcal{E}(x)$, $\mathcal{E}(\cdot)$ denotes the image encoder of VAE.

\noindent\textbf{Patch-to-Face Integration.}
The final goal of our method is to apply the generated adversarial patch to the target face image, creating a realistic adversarial sample capable of performing the impersonation attack. To achieve this, We first denoise the patch embedding to reconstruct in latent space:
\begin{equation}
p^{adv}= DDIM_{denoise}(p_m,m).
\end{equation}
% \begin{equation}
% p^{adv}=\underbrace{{DDIM_{denoise}}\circ\cdots\circ{DDIM_{denoise}}}_t(p_t).
% \end{equation}
where ${DDIM_{denoise}}$ denotes the diffusion denoising process, and $p^{adv}$ is the adversarial patch latent embedding.

% Specifically, this method uses a 2D image to capture the 3D coordinates of the facial point cloud, providing dense correspondences for each point in UV space. 

Then we decode the patch’s latent embedding into an adversarial patch image and leverage a UV location map~\cite{feng2018joint} for mask placement. 
The UV map captures the positional information of the 3D face and establishes dense correspondences between each point and its corresponding semantic meaning in the UV space, ensuring precise placement of the patch. By leveraging this map, we can digitally position the adversarial patch onto the face, resulting in a near-realistic adversarial example.
The process of the mask’s placement is as follows: given a face image, we first detect the landmark points to align the mask correctly with the face. The face image is then input into the 3D face reconstruction model for transforming the original image into the UV space. Subsequently, the adversarial patch is applied to the UV space face image, and the final image is reconstructed, producing a masked face image. 

Finally, we compute the cosine similarity between the adversarial sample and the target face image using the following loss function:
\begin{equation}
\label{eq: Ladv}
\mathcal{L}_{adv}=1-cos(f(\mathcal{R}_\theta(\mathcal{D}(p^{adv}),x)),f(\mathbf{x}^{tar})),
\end{equation}
where $\mathcal{R}_\theta$ is the rendering function responsible for applying the patch to the face, and $\mathcal{D}$ is the VAE decoder.

\tool~continuously refines the adversarial patch to achieve the impersonation attack by integrating these three optimized loss functions as follows:
\begin{equation}  
\arg\min_{p_m}\mathcal{L} = \lambda_{adv}\mathcal{L}_{adv} + \lambda_{atten}\mathcal{L}_{attn} + \lambda_{dir}\mathcal{L}_{dir},
\end{equation}
where $\lambda_{adv}$, $\lambda_{atten}$, and $\lambda_{dir}$ represent the weight factors of each loss. By default, we set these values as $\lambda_adv=10$, $\lambda_{atten}=10000$, $\lambda_{dir}=10$. The detailed optimization process is presented in Alg. \ref{alg}.

\begin{algorithm}[t]
\caption{\tool}
\begin{algorithmic}[1]
\Require Original face $x$, target face $x^{tar}$, iteration number $N$, pre-trained diffusion model $\mathrm{SD}$, loss function $\mathcal{L}_{attn}$, $\mathcal{L}_{dir}$, $\mathcal{L}_{adv}$, learning rate $\eta$.
\Ensure Optimized adversarial patch $p$ in pixel space.
\State Perform DDIM Inversion:
\[
x_{t} =DDIM_{inverse}(x, t),\] \[x^{tar}_{t} =DDIM_{inverse}(x^{tar}, t).\]
\State Initialize the adversarial patch:
\[
p_0= (1- \alpha) x_{t} + \alpha  x^{tar}_{t}.\]

\For{$m = 1$ to $N$}
    \State  Denoise the patch latent embedding:
    \[p_m \leftarrow {DDIM_{denoise}}(p_{m-1},1).\]
    \State Compute attention and direction loss $\mathcal{L}_{attn}$, $\mathcal{L}_{dir}$.
    \State Decode the patch latent embedding and generate adversarial face $x^{adv}$ using rendering function $\mathcal{R}_\theta$:
    \[x^{adv} = \mathcal{R}_\theta(\mathcal{D}(p_m),x).\]
    \State Compute attack loss $\mathcal{L}_{adv}$.
    \State Compute final loss \[\mathcal{L}=\lambda_{adv}\mathcal{L}_{adv} + \lambda_{atten}\mathcal{L}_{attn} + \lambda_{dir}\mathcal{L}_{dir}.\]
    \State Compute the gradient of the loss $g_m = \nabla_{p_m} \mathcal{L}$.
    \State Update the patch latent embedding: 
    \[p_m \leftarrow p_m - \eta g_m.\]
\EndFor
% \State \[p^{adv} \leftarrow p_N\]
\State \textbf{Finalize patch:} Decode the final latent embedding $p^{adv}=p_N$:
\[p=\mathcal{D}(p^{adv})\]

\State Return final adversarial patch $p$
\end{algorithmic}
\label{alg}
\end{algorithm}

\section{Experiments}
\label{sec:experiments}

\subsection{Experimental Setup}
\textbf{Datasets.} We conduct the experiments on LFW~\cite{huang2008labeled} and CelebA-HQ~\cite{karras2017progressive}, the two most popular benchmark datasets for low- and high-quality face images. Referring to the settings of~\cite{xiao2021improving}, we randomly choose 500 pairs of different identities from each dataset to measure the performance of the impersonation attack, where the images from the same pair are from different identities. 

\noindent\textbf{Models.} We evaluate on three face recognition models: FaceNet~\cite{schroff2015facenet}, CosFace~\cite{wang2018cosface}, and ArcFace~\cite{deng2019arcface}, all of which achieve over 99\% accuracy on the LFW validation set. When performing face recognition, the model extracts the feature representation of input face images and calculates the cosine similarity between the face image pairs. Then a threshold is applied to determine whether the pair represents the same identity. For each model, we select the threshold that achieves the highest accuracy on the LFW validation set.

\noindent\textbf{Metrics.} We use two metrics, attack success rate (ASR) and the number of queries (NQ), to evaluate the performance of impersonation attacks. ASR refers to the proportion of images that are successfully attacked in all test face images, where we ensure that the clean test images selected in the experiment can be correctly identified. We count the cases where the adversarial patch enables correct impersonation of the target identity. NQ refers to the number of queries to the target model required by the adversarial patch that can achieve a successful attack.

\noindent\textbf{Comparison.} We compare our method against five state-of-the-art adversarial patch techniques: GenAP~\cite{xiao2021improving}, Face3DAdv~\cite{yang2022controllable}, AT3D~\cite{yang2023towards}, RHDE~\cite{wei2022adversarial}, and Wei \textit{et al.}~\cite{wei2022simultaneously}.
GenAP, Face3DAdv, and AT3D are transfer-based methods that improve black-box attack success by leveraging surrogate models. GenAP optimizes adversarial patches on a low-dimensional manifold, while Face3DAdv uses a 3D generator for texture-based attacks. AT3D introduces adversarial 3D meshes with low-dimensional coefficient perturbations to enhance transferability across models.
RHDE and Wei \textit{et al.} focus on query-based attacks. RHDE uses a differential evolution algorithm to query patch locations, and Wei \textit{et al.} apply reinforcement learning to jointly optimize patch location and perturbation. 
% \delete{Both methods suffer from low query efficiency, limiting their practical effectiveness.}
% Implementation Details.

\begin{table*}[htbp]
\caption{Comparison results of the ASR (\%) and NQ between our method and other adversarial patch methods.}
\label{tab: comparison}
\renewcommand\arraystretch{1.0}
\centering 
\begin{threeparttable}
\resizebox{1\linewidth}{!}{
\begin{tabular}{lcccccc|cccccc}
\toprule[1.2pt]
\multirow{3}{*}{
\diagbox{Method}
% {dataset\yuec{need to revise}}
}         & \multicolumn{6}{c|}{LFW}                & \multicolumn{6}{c}{CelebA-HQ}          \\ \cline{2-13}
         & \multicolumn{2}{c}{ArcFace} & \multicolumn{2}{c}{CosFace}& \multicolumn{2}{c|}{FaceNet}   & 
         \multicolumn{2}{c}{ArcFace} & \multicolumn{2}{c}{CosFace} &\multicolumn{2}{c}{FaceNet}  \\ \cline{2-13} \cline{4-5}
         & {ASR}    & {NQ}        & {ASR}             & {NQ}   & {ASR}      & {NQ}      & {ASR}        & {NQ}& {ASR}        & {NQ}& {ASR}        & { NQ }        \\ \hline
GenAP~\cite{xiao2021improving}    & 53.50\%  &  -   & 50.00\%  &  -  & 46.50\% &  -   & 65.75\% &  -  & 58.25\% &  -   & 53.25\%  &  -   \\ 
Face3DAdv~\cite{yang2022controllable} & 40.06\%  &  -   & 33.23\%  &  -  & 46.65\% &  -   & 49.71\% &  -  & 36.61\% &  -   & 55.29\%  &  -   \\ 
AT3D~\cite{yang2023towards} & 50.50\%  &  -   & 41.00\%  &  -  & 45.20\% &  -   & 63.75\% &  -  & 59.00\% &  -   & 41.00\% &  -   \\
RHDE~\cite{wei2022adversarial} & 63.44\%  &  507   & 48.06\%  &  563  & 51.11\% &  636  & 56.10\% & 515  & 42.90\% & 653  & 48.18\% & 610  \\
Wei \textit{et al.}~\cite{wei2022simultaneously} & 49.50\%  &  36   & 40.08\% & 77 & 72.83\% & 27 & 44.48\% &  75 & 35.15\% & 99 & 72.06\% & 27 \\
\tool~ & \textbf{98.00\%}  &  \textbf{25}   & \textbf{93.20\%} & \textbf{47} & \textbf{95.00\%} & \textbf{25} & \textbf{96.40\%} & \textbf{44 } & \textbf{97.60\%} & \textbf{54} & \textbf{96.40\%} & \textbf{11} \\
\toprule[1.2pt]
\end{tabular}
}
% \begin{tablenotes}
% \item
% \end{tablenotes}
\end{threeparttable}
\end{table*}
\subsection{Experimental Results}
We present the experimental results of adversarial patches for query-based black-box impersonation attacks.
To prove the superiority, we compare our method with GenAP~\cite{xiao2021improving}, Face3DAdv~\cite{yang2022controllable}, AT3D~\cite{yang2023towards}, RHDE~\cite{wei2022adversarial}, and Wei \textit{et al.}~\cite{wei2022simultaneously}. For GenAP, Face3DAdv, and AT3D, we use their transferability to perform the black-box attack, their NQ is zero, so we use “-” to replace it. We followed the settings in their papers for each baseline, so the comparison is fair. Experiment results on two datasets for the three models are shown in Table~\ref{tab: comparison}.

\noindent \textbf{Effectiveness of \tool.}
From the experimental results, \tool~ demonstrates outstanding performance in both attack effectiveness and query efficiency. (1) Regarding attack effectiveness, our method achieves over 90\% attack success rate on both the LFW and CelebA-HQ datasets showing consistent effectiveness across different face recognition models such as ArcFace, CosFace, and FaceNet. Specifically, SAP-DIFF achieves a remarkable 98.00\% ASR on the LFW dataset for ArcFace and 97.60\% on the CelebA-HQ dataset for CosFace. (2) Regarding query efficiency, our method not only maintains a high success rate but also exhibits superior query efficiency. On the LFW dataset, our method requires an average of only 33 queries to complete the attack across these three models, and on the CelebA-HQ dataset, it needs an average of 37 queries. This query efficiency demonstrates that our method can perform efficient attacks with far fewer queries.

\noindent \textbf{Comparisons With SOTA Methods.}
We compare our method with existing approaches.
(1) Surrogate-based Models: Methods based on surrogate models rely on the transferability of the adversarial patches. GenAP, Face3DAdv, and AT3D typically show lower attack success rates, especially on the LFW dataset. For example, GenAP achieves ASR of 53.50\% on ArcFace, Face3DAdv reaches 40.06\%, and AT3D achieves 50.5\%. In contrast, our method achieves an outstanding 98.00\% success rate, significantly outperforming these methods. The trend continues on the CelebA-HQ dataset, where our method outperforms all surrogate-based approaches.
(2) Query-based Black-box Attack Methods: RHDE and Wei \textit{et al.} also exhibit relatively high attack success rates but still fall behind our method in both attack success rate and query efficiency. While their methods show competitive results, RHDE requires hundreds of queries, and Wei \textit{et al.} require a substantial number of queries as well. In comparison, our method achieves a higher attack success rate with fewer queries, demonstrating the effectiveness and efficiency of our approach.

In conclusion, our method significantly outperforms existing methods in both attack effectiveness and query efficiency due to its ability to leverage semantic feature extraction via diffusion models, allowing more effective and precise adversarial patch generation. Unlike traditional methods that rely on pixel-level perturbations, our approach targets high-level features, ensuring both higher efficiency and success rates in black-box impersonation attacks.

\vspace{-3pt}
\subsection{Universality Results}
Universality refers to the ability of an adversarial patch, originally generated for a specific target identity, to be effective across a diverse set of other face images. In simpler terms, a universal adversarial patch can consistently deceive face recognition models into misclassifying a variety of images as the target image, regardless of their original identity or the variations in the face images themselves.
To assess the universality of our method, we first randomly select a set of target images along with their corresponding generated adversarial patches. These patches are then applied to a pool of 500 different face images using the rendering function $\mathcal{R}_\theta$. This process simulates scenarios where the adversarial patch needs to work across a wide range of input images, not just the one for which it was originally designed. Following this, we test whether the adversarial patch is capable of causing these images to be misclassified as the target identity.

\begin{table}[t]
\caption{The universality performance, as measured by the attack success rate (ASR), of adversarial patches across different face recognition models on the LFW dataset.}
\label{tab: universality}
\begin{tabular}{c|ccc}
\hline
& Arcface & Cosface & Facenet \\ \hline
Universal ASR & 73.00\%    & 85.60\%  & 99.20\%  \\ \hline
\end{tabular}
\end{table}

Table~\ref{tab: universality} presents the universality of adversarial patches in different face recognition models (ArcFace, CosFace, and FaceNet) on the LFW dataset. The adversarial patches were applied to the face images using a rendering function, and the ability of these patches to cause misclassification to the target face image was tested. The results demonstrate that the universality of the adversarial patches varies across different models. The highest universality was achieved in FaceNet, with an attack success rate of 99.20\%, followed by CosFace at 85.60\%, and ArcFace at 73.00\%.
% The results demonstrate that the universality of the adversarial patches varies in different models, with FaceNet exhibiting the highest attack success rate (99.2\%), followed by CosFace (85.6\%) and ArcFace (73\%).

These findings underscore both the effectiveness and universality of the adversarial patches, demonstrating their potential as a robust attack method in different face recognition systems. The universality of our approach stems from the semantic structure of the adversarial patches, which are designed to target high-level features rather than pixel-specific perturbations. By focusing on the semantic features that represent the target identity, the patches are able to mislead a wide variety of face recognition models into misclassifying different face images. Furthermore, the attention disruption mechanism we employ ensures that the generated patches are not overly tailored to a specific identity, making them more adaptable and effective across a range of input face images. This combination of strategies enables our adversarial patches to achieve strong universality, ensuring their continued effectiveness across diverse face recognition systems.

\subsection{Ablation}
\begin{figure}[htbp]
  \centering
  \begin{subfigure}[b]{0.495\linewidth}
        \centering
        \includegraphics[width=\textwidth]{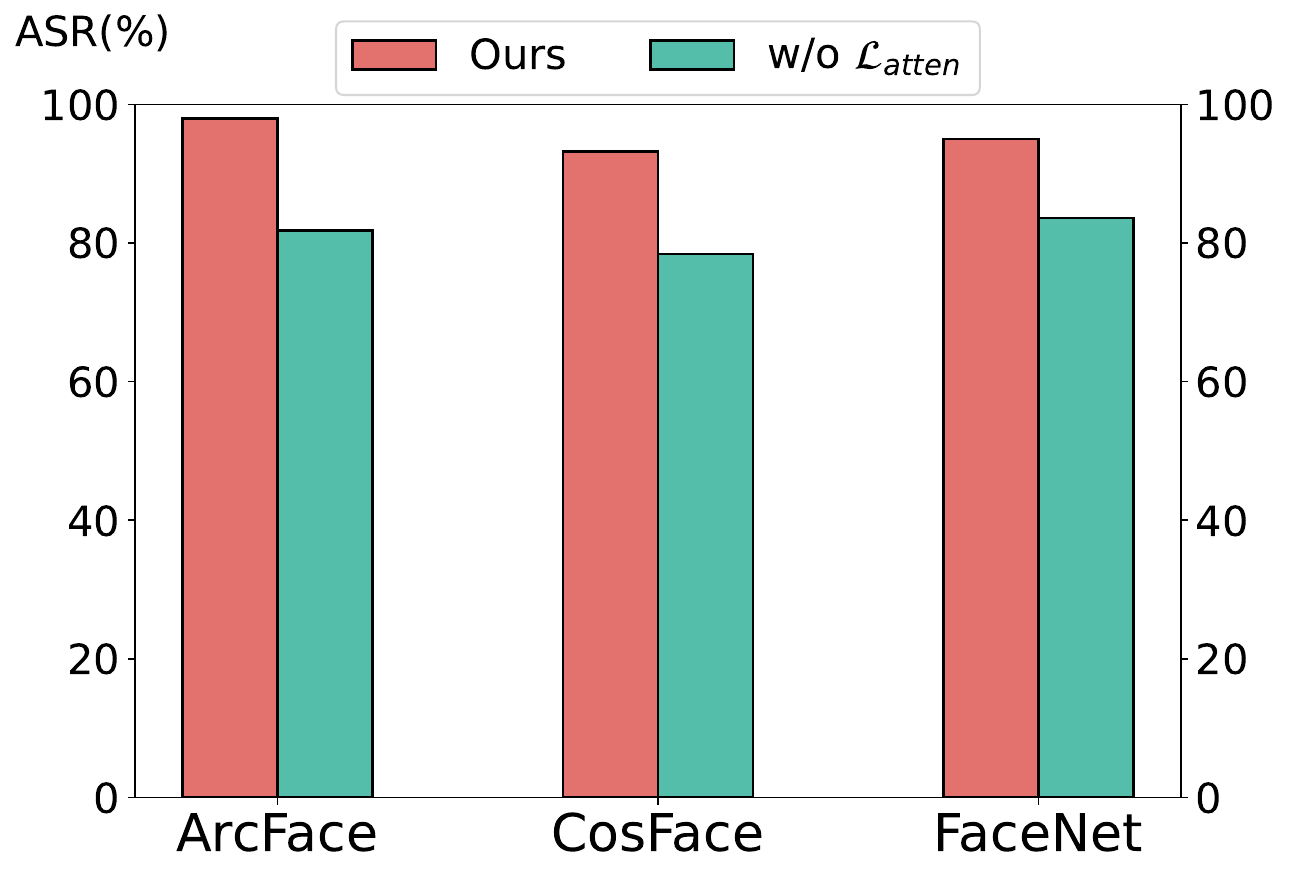}
        \caption{Impact of $\mathcal{L}_{atten}$ on ASR}
        \label{fig: ASR-atten}
  \end{subfigure}
  \hfill
  \begin{subfigure}[b]{0.495\linewidth}
        \centering
        \includegraphics[width=\textwidth]{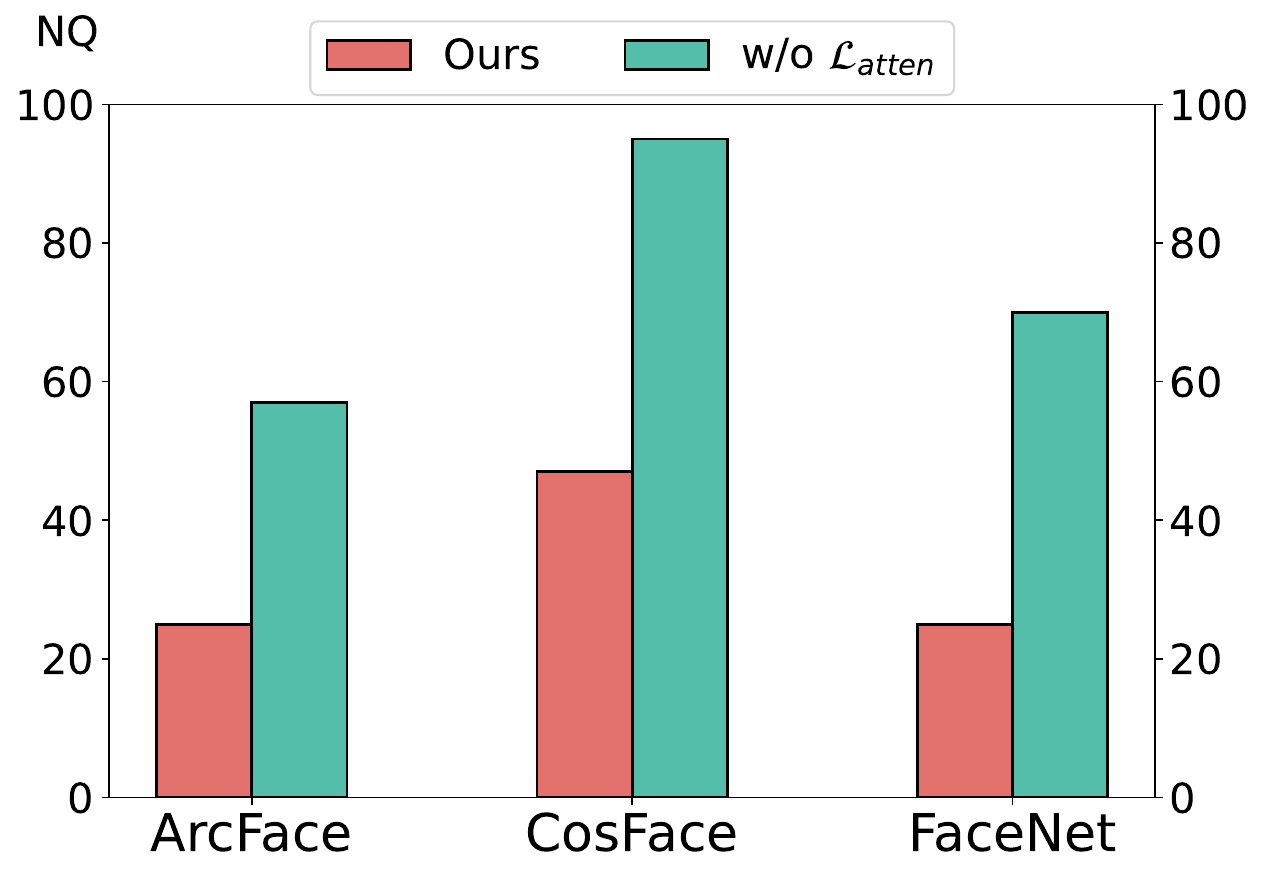}
        \caption{Impact of $\mathcal{L}_{atten}$ on NQ}
        \label{fig: NQ-atten}
  \end{subfigure}
  \caption{Ablation study of attention disruption loss.}
  \label{fig: impact of attention}
\end{figure}

\noindent \textbf{Attention Disruption}
To evaluate the impact of attention disruption, we conducted an ablation study by comparing the performance of the method with and without the attention disruption loss function. The results are shown in Fig.~\ref{fig: impact of attention}, which shows the ASR and  NQ for three different face recognition models: ArcFace, CosFace, and FaceNet, using the LFW dataset.
The results show a significant improvement in the ASR with the inclusion of the attention mechanism. Specifically, for ArcFace, ASR increases from 81.8\% to 98.0\%, for CosFace from 78.4\% to 93.2\%, and for FaceNet from 83.6\% to 95.0\%. This indicates that the attention disruption mechanism plays a critical role in enhancing the adversarial patch's ability to deceive the target models.
In terms of NQ, we observe a notable decrease when the attention mechanism is applied. For example, for ArcFace, NQ decreases from 56.52 to 24.99, for CosFace from 94.77 to 46.82, and for FaceNet from 69.54 to 24.83. The decrease in NQ suggests that the adversarial samples generated with the attention mechanism require fewer queries to achieve a successful attack, making the attack more efficient.

\begin{figure}[htbp]
  \centering
  \begin{subfigure}[b]{0.495\linewidth}
        \centering
        \includegraphics[width=\textwidth]{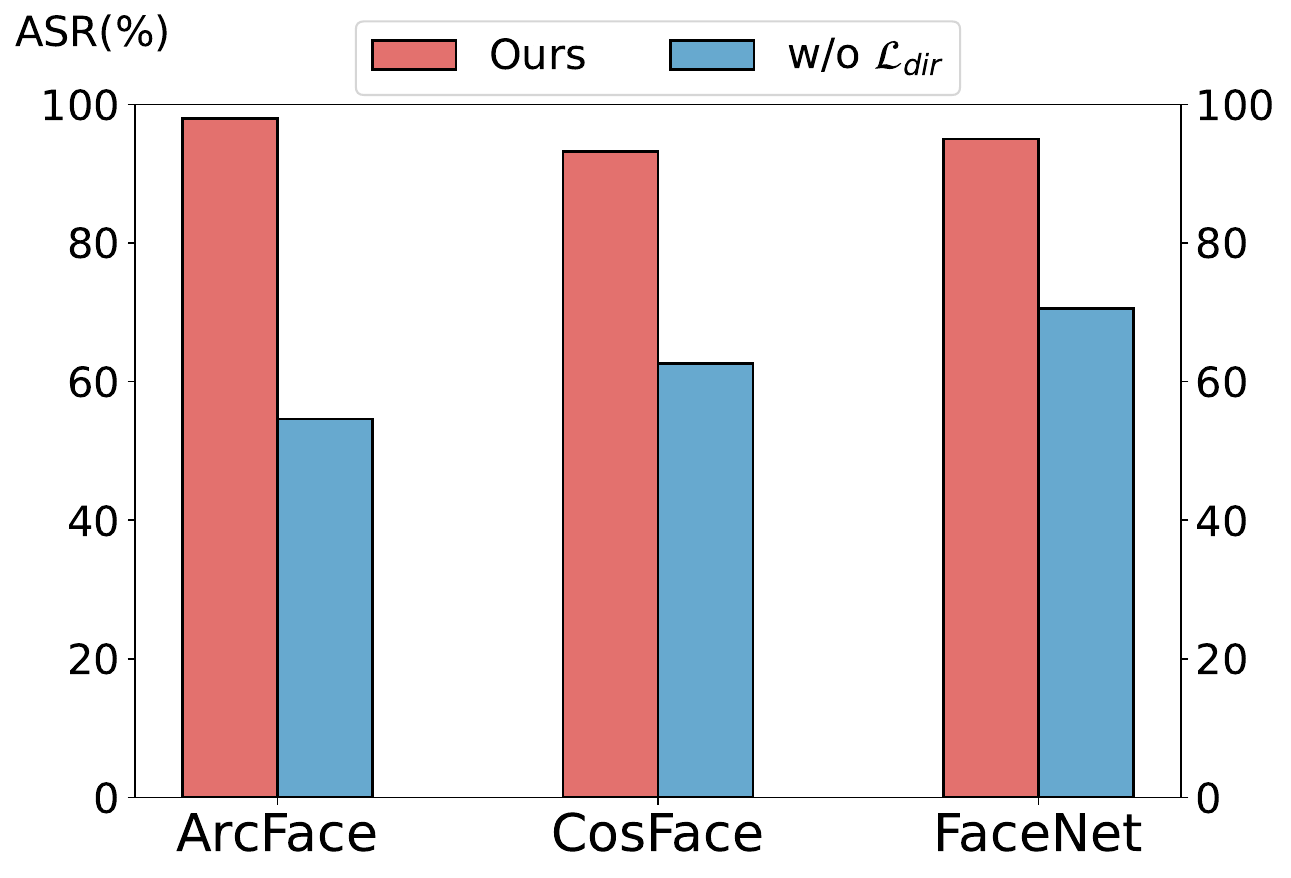}
        \caption{Impact of $\mathcal{L}_{dir}$ on ASR}
        \label{fig: ASR-dir}
  \end{subfigure}
  \hfill
  \begin{subfigure}[b]{0.495\linewidth}
        \centering
        \includegraphics[width=\textwidth]{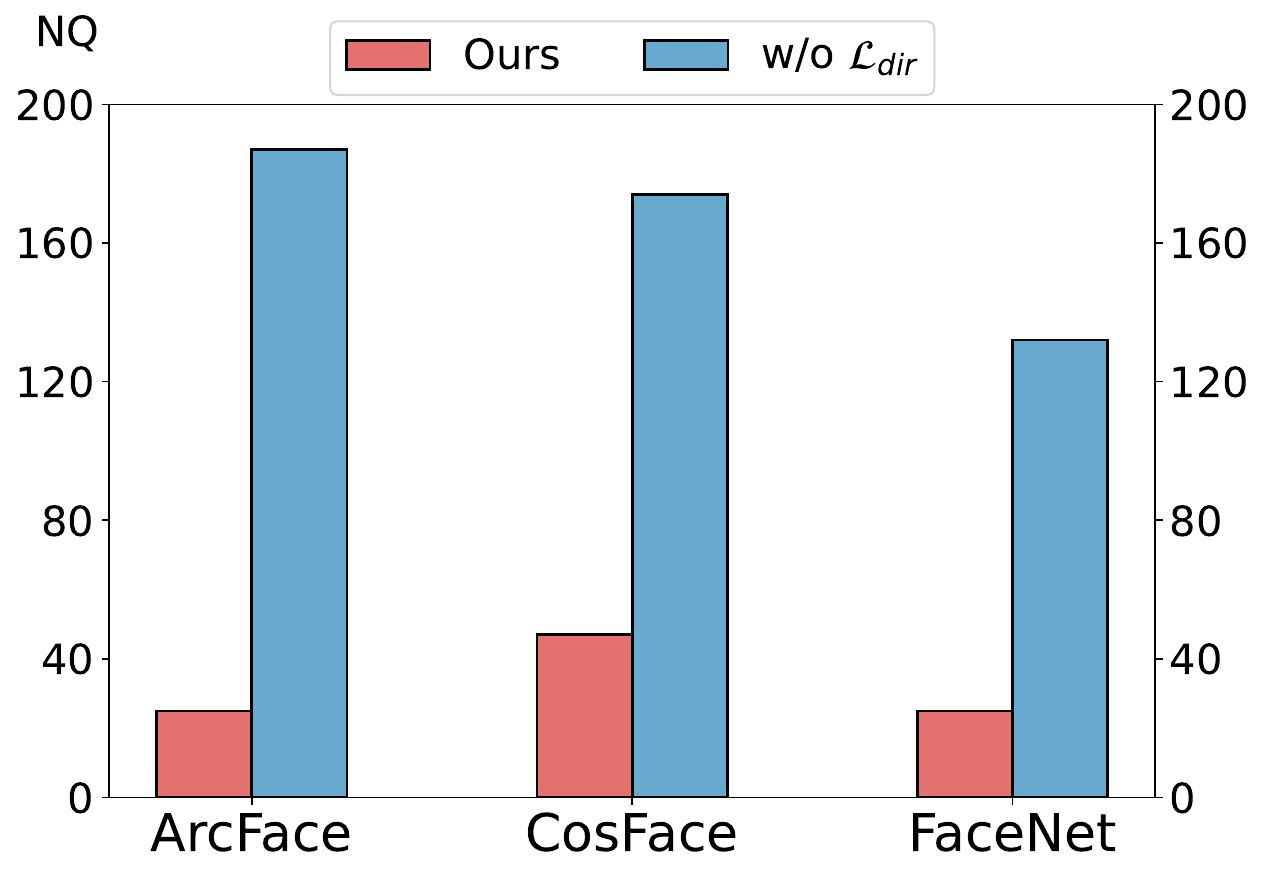}
        \caption{Impact of $\mathcal{L}_{dir}$ on NQ}
        \label{fig: NQ-dir}
  \end{subfigure}
  \caption{Ablation study of direction loss.}
  \label{fig: impact of direction}
\end{figure}

\noindent \textbf{Directional Guide}
To evaluate the impact of the directional loss, we compare the performance of the method with and without the directional loss function. The experiment is conducted on ArcFace, CosFace, and FaceNet using the LFW dataset. The results, shown in Fig.~\ref{fig: impact of direction}, present the ASR and NQ for each model.

The results demonstrate a significant decline in the attack success rate when the directional loss is removed. Specifically, for ArcFace, ASR decreases from 98.0\% to 54.6\%, for CosFace from 93.2\% to 62.6\%, and for FaceNet from 95\% to 70.6\%. This indicates that the directional loss plays a crucial and indispensable role in effectively guiding the adversarial patch toward the target face feature space. Without this loss, the adversarial attack becomes less effective. In terms of query number, for ArcFace, NQ increases from 24.99 to 187, for CosFace from 46.82 to 174, and for FaceNet from 24.83 to 132. This increase in NQ suggests that without the directional loss, the adversarial patches require more queries to successfully attack the model, thus reducing the attack's efficiency.

In conclusion, the ablation study highlights the significant impact of both attention disruption and directional loss in improving the adversarial attack's effectiveness and efficiency. When combined, these two components work synergistically to enhance the attack success rate while simultaneously reducing the query number, making the adversarial patch both more powerful and more efficient in query-based black-box settings.

The attention disruption and directional loss work synergistically to enhance the adversarial patch's effectiveness. The attention disruption diverts the model's focus from benign features to adversarial ones, while the directional loss steers the patch's optimization toward the target's feature space. This dual approach ensures the patch closely mimics the target identity, reducing the need for adjustments and minimizing queries for a successful attack. Together, they optimize both the patch's ability to deceive the model and its computational efficiency.

% The attention disruption forces the model to focus on the adversarial features, while the directional loss ensures that the patch is precisely aligned with the target's features.  The attention disruption works by effectively disrupting the model's focus on benign features, forcing it to concentrate more on the adversarial patch. On the other hand, the inclusion of the directional loss guides the optimization of the adversarial patch towards the feature space of the target face. This precise targeting ensures that the generated patch closely resembles the target identity, minimizing the need for further adjustments and thus reducing the number of queries required to successfully execute the attack. Together, these components improve the adversarial patch's ability to deceive the model while also optimizing its computational efficiency. 

\section{Discussion}

\textbf{Real-world Practice.} The success of adversarial attacks on commercial APIs hinges on the transferability of the generated adversarial samples. Our method, while achieving high specificity to the target model, faces limitations in transferability due to architectural and training data variations across different models. In practice, we attempt to use adversarial patches generated by a single FR model (FaceNet) to conduct adversarial attacks on three popular commercial FR APIs: Face++\footnote{https://www.faceplusplus.com}, Aliyun\footnote{https://vision.aliyun.com/facebody}, and iFLYTEK\footnote{https://global.xfyun.cn}, the ASR  are quite low: 3.5\% on Face++, 0\% on Aliyun, and 6\% on iFLYTEK.

To overcome this, we can employ ensemble learning~\cite{tramer2017ensemble, liu2016delving} incorporating multiple open-source models (FaceNet, ArcFace, and CosFace) during patch generation. This approach enhances cross-platform robustness by leveraging the diversity of model architectures. Specifically, we reformulate the adversarial loss as:
\begin{equation}
\mathcal{L}{adv}=1-\frac{1}{N}\sum_{i=1}^{N}cos(f_i(\mathcal{R}_\theta(\mathcal{D}(p^{adv}),x)),f_i(\mathbf{x}^{tar})),
\end{equation}
where $N$ is the number of models guiding the optimization process. 

We evaluate the ensemble learning method on the same three commercial APIs and achieve an ASR of 75.50\% on Face++, 5.50\% on Aliyun, and 19.50\% on iFLYTEK. The results indicate that ensemble learning can enhance the transferability of adversarial samples to a certain extent. Especially on the Face++ API, the ASR has significantly increased, which may be attributed to the similarity between the model architecture used in Face++ and those incorporated in ensemble learning. However, for commercial APIs that may employ proprietary model architectures not similar to our ensemble models, the ASR may still not be high. Expanding the diversity of the model ensemble could better approximate the characteristics of commercial systems, potentially increasing the success rates of impersonation attacks across more commercial APIs.

Moreover, our method leverages adversarial masks to attack face recognition models, which allows us to attack real-world deployed face recognition systems, potentially resulting in more severe consequences. For instance, in scenarios such as unlocking smartphones or gaining access to secure areas through face recognition, our adversarial patches could be printed and worn as masks, leading to misidentification and unauthorized access. The potential use of our method in real-world applications underscores the significant security vulnerabilities inherent in current face recognition systems, highlighting the urgent need for more resilient countermeasures and the development of defenses that can withstand such sophisticated adversarial attacks.

\textbf{Applicability.} Our method is designed to work with models that return output embeddings, which we use to compute the cosine similarity between the generated adversarial example and the target identity. This approach is highly effective when embeddings are available. However, it faces limitations when applied to commercial APIs or black-box models that only provide classification confidence scores. In these cases, we cannot directly apply our method to generate adversarial samples specifically for them. Nevertheless, as we discussed earlier, attacks based on transferability do not always produce optimal results and often lead to reduced effectiveness across different models or scenarios.

To address this challenge, we propose incorporating reinforcement learning~\cite{dong2019attention} into our framework in future work. We can design a reward function based on the model's confidence output, guiding the optimization of the adversarial patch toward the target identity. 
Reinforcement learning offers greater flexibility and adaptability and takes into account both the model’s response and the cosine similarity loss, enabling the method to perform well even in the absence of output embeddings. With this enhancement, we can target a wider range of black-box models and extend the applicability of our method to more real-world systems, providing a more dynamic solution for generating adversarial examples.

\section{Conclusion}
\label{sec:conclusion}
In this paper, we introduced a novel approach for generating adversarial patches to perform impersonation attacks on face recognition (FR) models, leveraging diffusion models for latent space manipulation. Our method incorporates an optimization strategy that includes attention disruption to prevent the model from generating benign features, directional loss to guide the patch toward the target identity’s feature space, and a UV location map to ensure the patches are applied realistically on the face. This combination ensures that the generated patches are both semantically effective and imperceptible, effectively deceiving face recognition systems through query-based black-box attacks, without requiring knowledge of the target model's internal parameters. Extensive experiments conducted on benchmark datasets across popular FR models highlight the good performance of our method.

% \begin{acks}
% \end{acks}

% \clearpage
%% References
\bibliographystyle{ACM-Reference-Format}
\bibliography{reference} 

\end{document}